\documentclass[letterpaper, 10 pt, conference]{ieeeconf}  

\IEEEoverridecommandlockouts                              

\usepackage{algorithm}
\usepackage{algorithmic}
\usepackage{amsmath} 
\usepackage{amssymb}  
\usepackage{graphicx} 
\usepackage{amsfonts}
\usepackage{xcolor}
\usepackage{cite}
\usepackage{units}
\usepackage{hyperref}

\newcommand{\executeiffilenewer}[3]{\ifnum\pdfstrcmp{\pdffilemoddate{#1}}{\pdffilemoddate{#2}}>0{\immediate\write18{#3}}\fi}
 
\newcommand{\q}{\boldsymbol{q}}
\renewcommand{\r}{\boldsymbol{r}}

\newcommand{\A}{\boldsymbol{A}}
\newcommand{\B}{\boldsymbol{B}}

\newcommand{\R}{\boldsymbol{R}}

\newcommand{\f}{\boldsymbol{f}}
\newcommand{\x}{\boldsymbol{x}}
\newcommand{\y}{\boldsymbol{y}}

\renewcommand{\u}{\boldsymbol{u}}

\newcommand{\h}{\boldsymbol{h}}

\newcommand{\Q}{\boldsymbol{Q}}

\newcommand{\g}{\boldsymbol{g}}

\newcommand{\ve}{\boldsymbol{v}}
\newcommand{\0}{\boldsymbol{0}}

\newcommand{\M}{\boldsymbol{M}}

\newcommand{\C}{\boldsymbol{C}}

\usepackage{amsmath} 
\usepackage{amssymb}  
\usepackage{amsfonts}
\usepackage{subfig}
\usepackage{booktabs}

\usepackage{lipsum}

\usepackage{ulem}
\usepackage{pgffor}

 \usepackage[outline]{contour}

\usepackage{xcolor, url}
\usepackage{graphicx}

\DeclareMathAlphabet{\pazocal}{OMS}{zplm}{m}{n}

\title{\LARGE \bf Haptic Teleoperation of High-dimensional Robotic Systems Using a Feedback MPC Framework}

\author{Jin Cheng, Firas Abi-Farraj, Farbod Farshidian, Marco Hutter
\thanks{The research leading to these results has received funding from armasuisse Science and Technology, the Swiss National Science Foundation through the National Centre of Competence in Research Robotics (NCCR Robotics), and ANYmal Research, a community to advance legged robotics.
Authors are with the Robotic Systems Lab, ETH Zurich. Email: {\tt\small \{jicheng, fabifarraj, farbodf, mahutter\}@ethz.ch}}
}

\begin{document}

\maketitle
\thispagestyle{empty}

\begin{abstract}
Model Predictive Control (MPC) schemes have proven their efficiency in controlling high degree-of-freedom (DoF) complex robotic systems. However, they come at a high computational cost and an update rate of about tens of hertz. This relatively slow update rate hinders the possibility of stable haptic teleoperation of such systems since the slow feedback loops can cause instabilities and loss of transparency to the operator. This work presents a novel framework for transparent teleoperation of MPC-controlled complex robotic systems. In particular, we employ a feedback MPC approach \cite{grandia2019feedback} and exploit its structure to account for the operator input at a fast rate which is independent of the update rate of the MPC loop itself. We demonstrate our framework on a mobile manipulator platform and show that it significantly improves haptic teleoperation's transparency and stability. We also highlight that the proposed feedback structure is constraint satisfactory and does not violate any constraints defined in the optimal control problem. To the best of our knowledge, this work is the first realization of the bilateral teleoperation of a legged manipulator using a whole-body MPC framework.
\end{abstract}

\section{Introduction}
\label{sec:introduction}
Autonomous robots have shown significant improvements over the past years, but their capabilities are still limited, especially when interacting with unknown and unstructured environments. The need for human intervention is, therefore, still essential. Teleoperation allows humans to act, through a robotic avatar, in a remote (deep ocean, space) or dangerous (nuclear, chemical) environment while ensuring their own safety. The need for mobile manipulators which can be teleoperated is therefore vital, and the lack of such systems was, for example, flagrantly highlighted during the Fukushima nuclear disaster. Recent developments on quadrupedal robots have shown excellent and robust locomotion capabilities across different types of terrain making them suitable for acting in urban environments such as stairs, doors, or any other untraversable hurdles for wheeled-base systems. Model Predictive Control (MPC) has shown great performance in the locomotion and manipulation tasks of legged manipulators. However, the computational burden and slow update rate of MPC are the major challenge of bilateral teleoperation on such systems. This slow update rate leads to a significant delay between the current target of MPC and the actual command from the operator which leads to a loss of transparency and instabilities in the haptic feedback loop. 

This work proposes a novel haptic teleoperation framework for complex high-DoF robotic systems controlled by a whole-body MPC. The main idea is to exploit the feedback MPC \cite{grandia2019feedback} and make the policy target-dependent such that it can be evaluated using a new target at a high rate. To this end, we first augment the MPC's state with the desired position and velocity, with which a DDP-based MPC solver can predict the command and produce a feedback policy on the target. Second, we employ a similar feedback MPC framework as \cite{grandia2019feedback} where the optimized feedback policy of the MPC is used to evaluate the plan at a higher rate in the whole-body controller. This target-dependent policy, in turn, allows executing human commands at a fast rate independently from the update rate of the MPC itself, resulting in a more transparent and stable teleoperation loop. 

\begin{figure}[t]
\centering
\includegraphics[width=1.0\columnwidth]{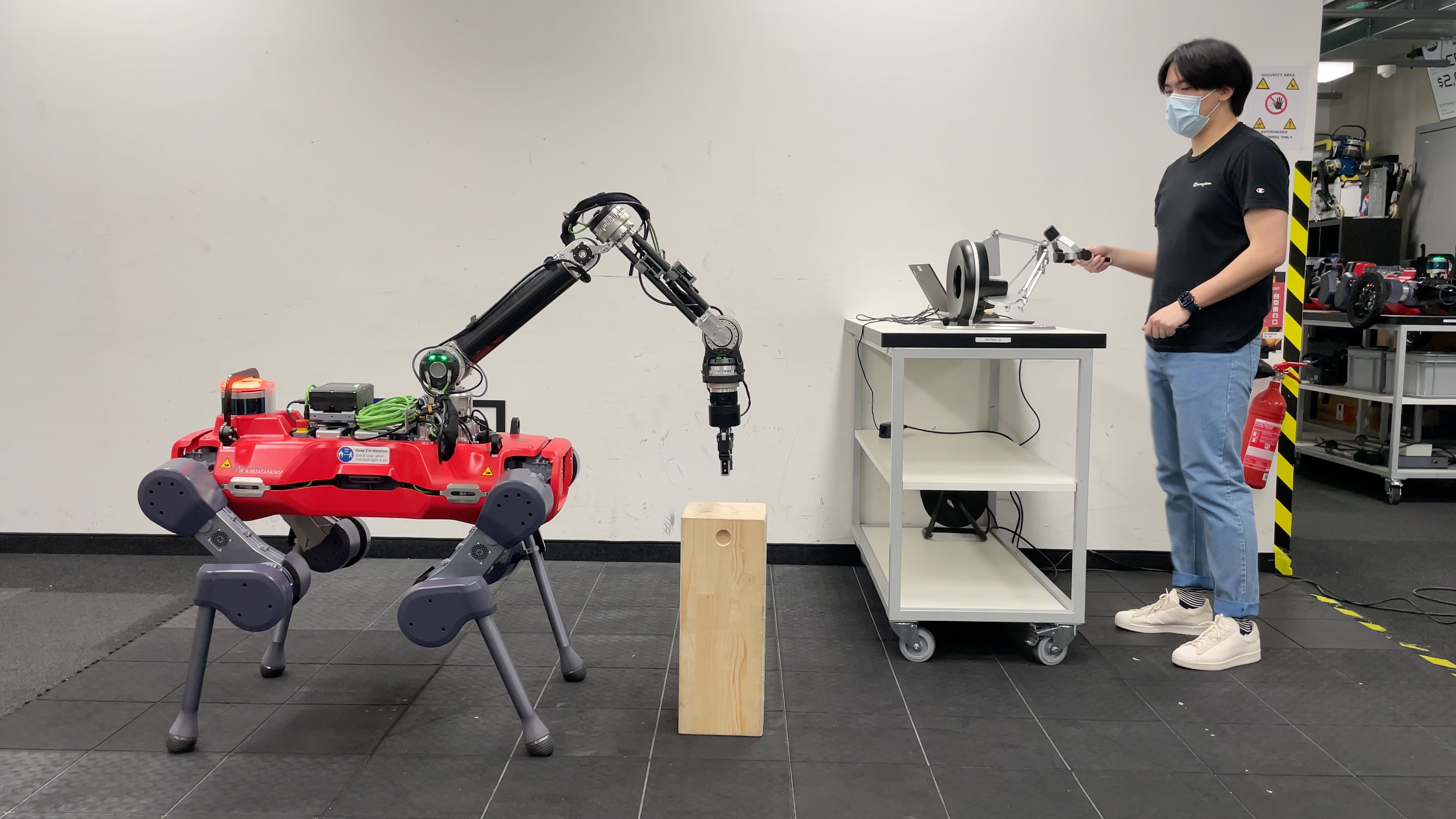}
\caption{Experimental setup showing the quadrupedal manipulator \textit{(left)} and the operator acting on the haptic device \textit{(right)}.}
\label{fig:experiment}
\vspace{-5mm}
\end{figure}

Without loss of generality, we focus on teleoperating the end-effector position with a fixed desired orientation on a quadrupedal manipulator (Fig.~\ref{fig:experiment}), but this framework can be applied to any MPC-controlled robot. Finally, we validate our framework in simulation and on hardware in a bilateral teleoperation setup demonstrating that the delay and tracking errors can be significantly reduced, the force feedback is more stable and that the method is constraint satisfactory.

The contributions of this work are as follows:
\begin{itemize}
    \item We show one of the first deployments of a whole-body MPC for teleoperation on a legged mobile manipulator to the best of our knowledge.
    \item We present a novel framework to handle the delay introduced by the slow update rate of the MPC to achieve a stable and transparent bilateral teleoperation.
    \item We validate our framework both in simulation and on hardware for teleoperating a quadrupedal manipulator. 
\end{itemize}


\section{Related Work}
\label{sec:relatedwork}

\subsection{MPC for Quadrupedal Manipulator}
\label{subsec:modelpredictivecontrol}
MPC has shown its capability of controlling complex dynamic systems but is typically computationally demanding. Variants such as tube MPC \cite{langson2004robust} can optimize an affine structured policy with a predefined feedback gain. When the new policy is not available, this scheme can be evaluated at a higher rate. But instead of using a fixed feedback gain, controllers using Dynamic Programming (DP) such as Linear Quadratic Regulator (LQR) with time-varying feedback gains can also be formulated in a receding horizon fashion. For nonlinear systems, variants of the DP-based method such as Differential Dynamic Programming (DDP) \cite{mayne1966second}, \cite{jacobson1968new} have been proposed to iteratively optimize a sequence of control input around the nominal trajectory rolled out by the latest nominal input sequence. Two efficient variant of DDP methods are known as iterative LQR (iLQR) \cite{li2004iterative}, \cite{todorov2005generalized} for discrete-time system and sequential linear quadratic (SLQ) method \cite{farshidian2017sequential} for continuous-time system, which solve the extremal problem sequentially based on the latest linearization of the dynamics and quadratic approximation of the cost. 

For legged robots, a variant of the SLQ algorithm was presented and formulated in an MPC scheme with multiple threads to achieve real-time motion planning \cite{farshidian2017efficient}, \cite{farshidian2017real}. Using the SLQ feedback policy in an MPC fashion was deployed and validated on hardware in \cite{grandia2019feedback}, where a hierarchical inverse dynamics controller is integrated into the whole-body controller to process the policy from the MPC \cite{bellicoso2016perception}. 

For quadrupedal manipulators, offline planned tasks such as lifting and throwing can be achieved using a separated arm controller \cite{murphy2012high} in a similar control framework as wheeled manipulators \cite{kuindersma2016optimization}, \cite{rehman2016towards}. However, to achieve adaptation of the base for manipulation tasks, the controllers for locomotion and manipulation on humanoids and quadrupedal manipulators are usually unified as one. In \cite{bellicoso2019alma}, the end-effector tracking task was combined with the hierarchical whole-body controller to achieve torso adjustment if the gripper is out of kinematic range. A unified MPC framework has been proposed \cite{sleiman2021unified}, \cite{sleiman2021constraint} by formulating the locomotion and manipulation tasks as one unified multi-contact optimal control problem using the SLQ-MPC planner. The framework can be implemented onboard in real-time and achieve good performance in tasks such as tracking and pushing.

\subsection{Teleoperation on Legged Robots}
\label{subsec:teleoperationonleggedrobots}
Several attempts towards teleoperating legged systems have been presented in literature. In \cite{o2013humanoid}, the humanoid HUBO was teleoperated using a motion-tracking device but no haptic feedback was provided to the operator. In \cite{zucker2015general}, the DRC-HUBO humanoid robot was also controlled via a visual interface without any haptic feedback. In \cite{neo2007whole} and \cite{stilman2008humanoid}, the authors used a single 3-DoF joystick to give targets for specific parts of the humanoid body, while letting the robot balance itself but still with no haptic feedback from the robot.
A haptic teleoperation framework for teleoperating a humanoid robot was presented in \cite{evrard2009intercontinental} where the authors used an inverse dynamics controller to control the robot end-effector. While the results presented in the paper are interesting, the controller used does not exhibit the advantages of an MPC scheme which remains more robust in real-world scenarios. In \cite{brygo2014tele} and \cite{abi2018humanoid}, the haptic feedback provided the operator with information about the balance and stability of a humanoid robot, but the interaction force between the robot and the environment was not fed back to the user. In \cite{risiglione2021passivity}, a passivity-based control structure was proposed for the stable bilateral teleoperation of the arm of an MPC-controlled legged manipulator. While stable teleoperation was achieved, the slow update rate of the controller (running at 70 Hz) had a significant impact on the quality and transparency of the teleoperation experience itself.

\section{Preliminaries}
\label{sec:preliminaries}

In this section, we will introduce some concepts from existing literature which are needed for understanding the proposed method. We also present a summary of the controller used on the robot and its different components presented previously in \cite{grandia2019feedback} and \cite{sleiman2021unified}.

\subsection{MPC Planner}
\label{subsec:mpcplanner}
We formulate the planning task as a nonlinear constrained optimal control problem:
\begin{align}
\min_{u(\cdot)}\, & \phi(\x(T)) + \int_0^T l(\x(t), \u(t), t)\,dt\\
s.t. \, &\dot{\x}(t) = \f(\x(t), \u(t), t) \\
&\x(0) = \x_0 \\
&\h_{eq}(\x(t), \u(t), t) = 0 \label{constriant:equality}\\
&\h_{in}(\x(t), \u(t), t) \geq 0, \label{constraint:inequality}
\end{align} 
where $\x(t)\in \mathbb{R}^{n_x}$ is the state of the system, $\u(t) \in \mathbb{R}^{n_u}$ is the control input at time $t$. $l$ is a time-varying scalar stage cost, and $\phi$ is a terminal scalar cost at the state $\x(T)$. Using the SLQ-MPC framework which runs the constrained SLQ algorithm in a receding horizon fashion \cite{farshidian2017sequential}, \cite{farshidian2017efficient}, \cite{farshidian2017real}, a feedback structured policy consisting of a sequence of feedforward inputs and feedback gains that minimizes the cost can be calculated.

\subsubsection{System dynamics}
A legged manipulator is a rigid body system with an unactuated floating base and several fully actuated limbs (including legs and arms). 
We model this system through a full centroidal dynamics approach \cite{sleiman2021unified}. The centroidal dynamics part of this model describes the projected dynamics of the robot at its Center of Mass (CoM) as
\begin{equation}
    \dot{\h}_{com} = 
    \begin{bmatrix}
    \sum_{i=1}^{n_c} \f_{c_i} + m\g \\
    \sum_{i=1}^{n_c} \r_{com,c_i}\times \f_{c_i}
    \end{bmatrix} \label{eq:centroidaldynamics},
\end{equation}
where $\h_{com}$ is the centroidal momentum. $\r_{com,c_i}$ is the position of contact point $c_i$ relative to the CoM, $\f_{c_i}$ is the external contact force from the environment, and $m \g$ is the gravitational force acting on CoM. The rate of change in the generalized coordinates can affect the centroidal momentum through 
\begin{equation}
    \label{eq:centroidal_momentum}
    \h_{com} = \A_b(\q)\dot{\q}_b + \A_j(\q)\dot{\q}_j,
\end{equation}
where $\A(\q) = [\A_b(\q), \A_j(\q)]\in \mathbb{R}^{6 \times (6+n_a)}$ is the centroidal momentum matrix. $\q_b = (\r_{IB}, \boldsymbol{\Phi}_{IB}^{zyx})\in \mathbb{R}^6$ is the base pose relative to the inertial frame (world frame) and $\q_j \in \mathbb{R}^{n_a}$ is the joint velocities of $n_a$ actuated joints. Equation \eqref{eq:centroidal_momentum} can be rearranged as
\begin{equation}
    \dot{\q}_b = \A_b^{-1}(\h_{com} - \A_j\dot{\q}_j) \label{eq:basecoordynamics}, 
\end{equation}
which defines the rate of base coordinates in terms of the centroidal momentum and the joint velocities. Moreover, the dynamics of the actuated joint coordinates $\q_j$ can be described as
\begin{equation}
    \dot{\q}_j = \ve_j \label{eq:jointdynamics},
\end{equation} 
where $\ve_j \in \mathbb{R}^{n_a}$ is the joint velocities.
By defining the state vector as $\x = (\h_{com}, \q_b, \q_j)\in \mathbb{R}^{12+n_a}$, and the input vector as $\u = (\f_{c_1},..., \f_{c_{n_c}}, \ve_j)\in\mathbb{R}^{3n_c+n_a}$, the flow map of dynamic system $\dot{\x} = f(\x, \u)$ can be defined as the collection of equations \eqref{eq:centroidaldynamics}, \eqref{eq:basecoordynamics}, \eqref{eq:jointdynamics}. 

\subsubsection{Constraints} 
The active contact points depend on the predefined gait schedule. The gait-dependant constraints of the system include zero-velocity of contact feet, zero contact force of swing feet. Also, when a foot contact point is swinging, it should track a generated reference trajectory in the direction of the normal vector of the local surface to avoid foot scuffing. To ensure dynamical feasibility, we also need to enforce the contact forces on the feet to remain in the friction cone and non-pulling. The handling of constraints in the SLQ-MPC is described in detail in \cite{sleiman2021constraint}.

\subsubsection{Cost function} 
The following cost function is defined in the MPC planner for the end-effector position tracking.
\begin{equation}
\begin{split}
    l(\x, \u) = \|\mathrm{\mathrm{\Delta}} \x_{ee, p} \|^2_{\Q_{ee,p}} \!\!\! + \| \mathrm{\Delta} \x_{ee, o}\!  \|^2_{\Q_{ee,o}} \!\!\! + \| \mathrm{\Delta} \u\|^2_{\R},
\end{split}
\end{equation}
where $\R$ is a positive definite matrix, and $\Q_{ee,p}$, $\Q_{ee,o}$ are positive semi-definite weighting matrices. The vectors $\rm{\mathrm{\Delta}} \x_{ee, p}$, $\rm{\mathrm{\Delta}} \x_{ee, o}$ and $\rm{\mathrm{\Delta}} \u$ correspond to the error in end-effector position, error in end-effector orientation, and the difference in input respectively. The error in the end-effector position is calculated by
\begin{equation}
    \mathrm{\mathrm{\Delta}} \x_{ee, p} = \x_{ee, p} - \x_d, \label{eq:trackingerror}
\end{equation}
where $\x_{ee, p}$ is the end-effector position calculated from forward kinematics, and $\x_d$ is the desired position from the desired discredited trajectory over the MPC horizon, which is defined in
\begin{equation}
    \y_d = (\x_{d, t_0}, ..., \x_{d, t_k}),
    \label{eq:mpc_input}
\end{equation}
where $\y_d \in \mathbb{R}^{3 \times (k+1)}$ and $k \in  \mathbb{R}$ is the discretization resolution. $\y_d$ is defined differently according to the method used and will be described in more details in \ref{subsec:desiredposandvelasaugmentedstates} and \ref{sec:results}.

\subsection{Whole-body Controller}
\label{subsec:wholebodycontroller}
A time-varying, state-affine feedback policy can be calculated based on the quadratic approximation of the cost function and the linear approximation of the dynamics and constraints. The feedback policy takes the form
\begin{equation}
\begin{split}
\pi(\x(t), t) &= \Bar{\u}(t) + K(t)(\x(t) - \Bar{\x}(t)) \\
&= \u_{ff}(t) + K(t)\x(t),
\end{split}
\label{eq:mpcfeedback}
\end{equation}
where $\Bar{\u}(t)$ and $\Bar{\x}(t)$ represent the optimal input sequence and the optimal state sequence. Although feedback policy with full states can be evaluated with interpolated feedforward term and feedback gain, it requires techniques such as frequency shaping to implement on hardware. Otherwise, instability issues can occur due to the negligence of the dynamics of the low-level controller in the high-level control design \cite{grandia2019feedback}. In our work, we will not use the feedback policy on the full states. This will be discussed in more details in Sec.~\ref{subsec:desiredposandvelasaugmentedstates}.

\begin{figure}[t]
   \centering
   \includegraphics[trim=40 0 70 0,clip,width=0.9\linewidth]{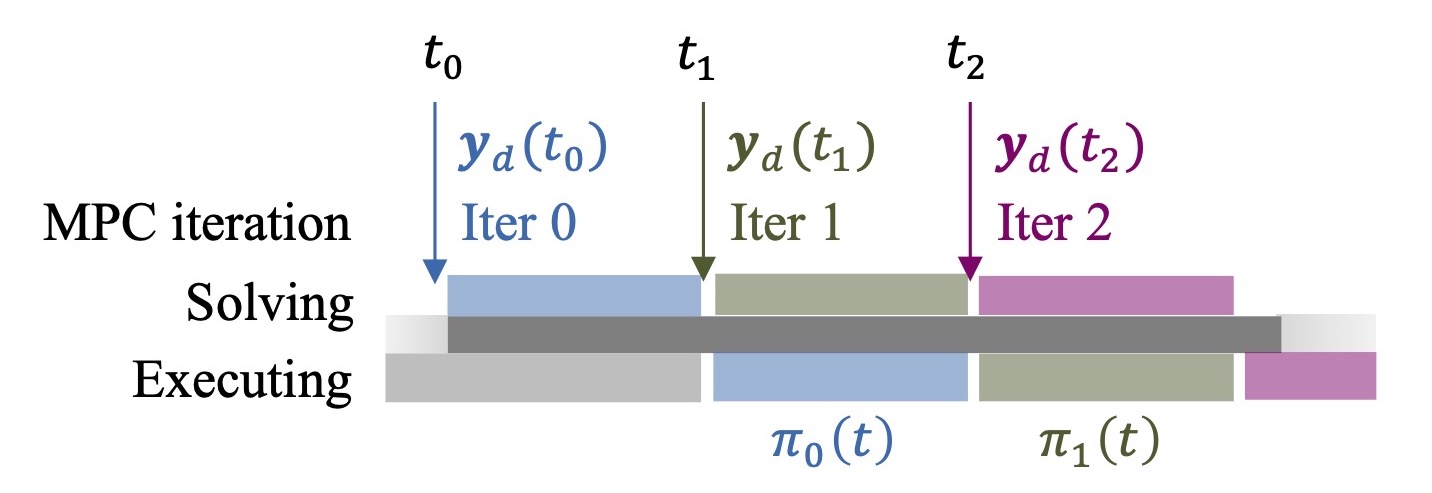}
   \caption{MPC planner might fail in teleoperation. $\y_d$ is the target trajectory from the operator, $t_0, t_1, t_2$ are the time point when a new MPC policy is available and switched into use, $\pi_0, \pi_1$ are the policy calculated from iteration 0 and iteration 1 respectively.}
   \label{fig:telefail}
   \vspace{-5mm}
\end{figure}

\begin{figure*}[t]
   \centering
   \includegraphics[width=0.85\textwidth]{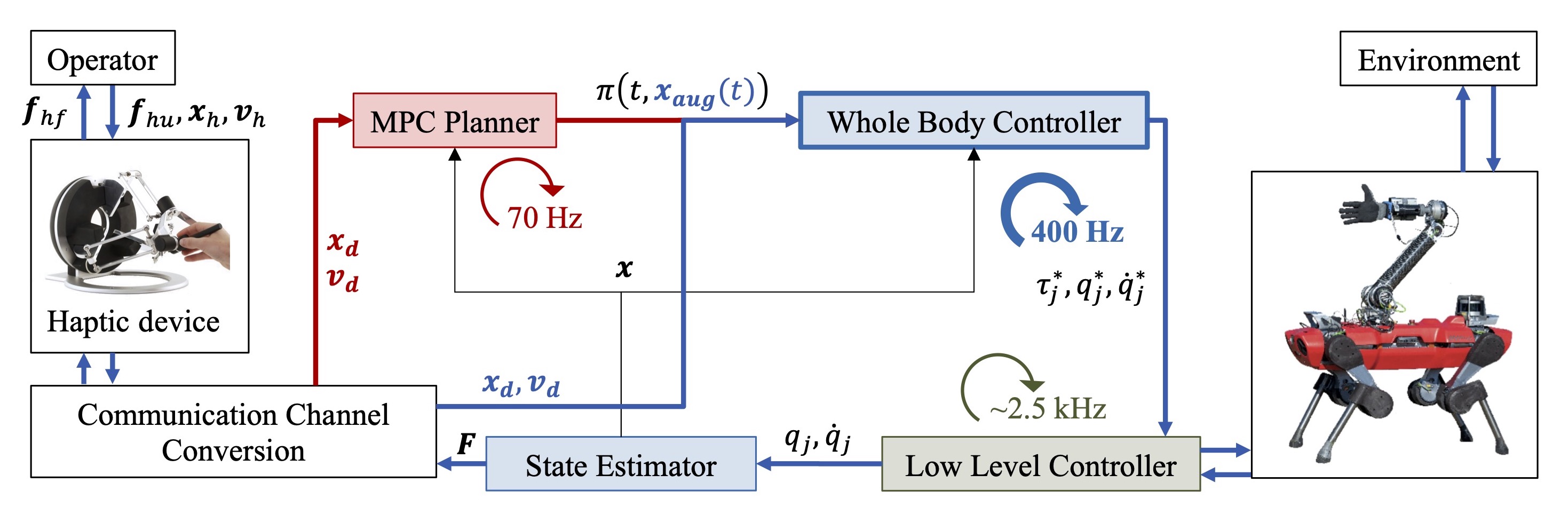}
   \caption{The teleoperation pipeline using the proposed method. Operator commands are used to update the MPC policy at every iteration of the whole-body controller allowing for a much faster bilateral teleoperation loop.}
   \label{fig:ourtelestructure}
   \vspace{-5mm}
\end{figure*}

\section{Method}
\label{sec:method}
The delay due to the MPC planner is demonstrated in Fig.~\ref{fig:telefail}. Here, the MPC is inquired at successive time instances, $t_0, t_1, t_2$. The frequency of these calls depends on the dimension of the problem and the available computational resources, which is about 12-15 ms per call in our case. MPC optimizes its plan at each call based on the latest state measurement and the user-defined desired target. This plan is then passed to the whole-body controller, where it tracks the plan until a new MPC output is available. However, if the user commands a new desired trajectory within one call of MPC, that won't be immediately taken into account since the MPC is already triggered with an old command, and the whole-body controller is only informed about the new command after the completion of the next MPC call. In the case of complex systems such as a legged manipulator, this loop runs at a rate of 70~Hz, introducing a significant delay in the teleoperation control loop, which may render it unstable and deteriorate the teleoperation transparency.

The proposed framework is illustrated in Fig.~\ref{fig:ourtelestructure}. The operator acts on a haptic device to command the end-effector of a legged manipulator. The desired position and velocity are sent into the MPC planner, which generates a feedback policy at 70 Hz. This policy will be evaluated using the new commands in the whole-body controller at a rate of 400 Hz. Then the low-level module can generate the command torque using a PD controller, and the actual joint position and velocity are taken from the state estimator to provide feedback to the MPC planner and whole-body controller. Finally, the readings of a force-torque sensor are fed back to the operator as haptic feedback to close the bilateral haptic teleoperation loop.

\subsection{System Overview}
\label{subsec:systemoverview}
We consider a classical bilateral teleoperation setup where the operator acts on a 3-DoF haptic device to control the position of the end-effector of an $n$-DoF legged manipulator. The operator commands the target position of the end-effector by acting on the haptic device, and receives haptic feedback informing about the robot interaction forces with the environment. A force-torque sensor mounted on the end-effector of the robot arm is used to estimate the interaction force and feed it back to the operator.

Let us define three reference frames: $\pazocal{F}_h \in SE(3)$ attached to the base of the haptic device, $\pazocal{F}_{r} \in SE(3)$, a frame which its origin is at the projection of the center of mass of the robot base on the ground and is oriented along the robot yaw direction (with the vertical looking up) and $\pazocal{F}_w \in SE(3)$ as a fixed world frame.

The configuration of the haptic interface and its Cartesian velocity are defined in  $\pazocal{F}_{w}$ by $\x_h\in\mathbb{R}^3$ and $\ve_h\in\mathbb{R}^3$, respectively. The device is assumed to be a 3-DoF device and is modelled as a gravity-pre-compensated generic mechanical system,
\begin{equation}\label{eq:master_device}
\M(\x_h)\dot{\ve}_h +\C(\x_h,\,\ve_h)\ve_h =\f_{hf} + \f_{hu} + \B \ve_h,
\end{equation}
where $\M(\x_h)\in\mathbb{R}^{3\times 3}$ is the positive-definite symmetric inertia matrix, $\C(\x_h,\,\ve_h)\in\mathbb{R}^{3\times 3}$ are the Coriolis and centrifugal terms, $\f_{hf},\,\f_{hu}\in\mathbb{R}^3$ are the  haptic feedback and the human operator's forces, respectively, and $\B\in\mathbb{R}^{3\times 3}$ is a damping matrix for stabilizing the system.
 
We define the desired linear velocity of the end-effector of the robot arm in $\pazocal{F}_{r}$ as $^r\ve_d \in\mathbb{R}^{3}$ and its desired position in the same reference frame as $^r\x_d \in\mathbb{R}^{3}$. A position-to-position coupling between the haptic device and the remote system is then defined by setting 
\begin{equation}
\begin{cases}
^r\x_d=~^r\x_{d,0} + (\x_h-\x_{h,0})  \\
^r\ve_{d}=\ve_h,
\end{cases}
\end{equation}
where $\x_{h,0}$ and $^r\x_{d,0}$ are the initial positions of the haptic device and the robot end-effector respectively. A clutching mechanism is implemented to allow the user to freely move the haptic device without moving the robot if needed. Before clutching back in, $\x_{h,0}$ and $^r\x_{d,0}$ are updated to the last known positions of the haptic device and the robot end-effector respectively.

The desired robot target position, $\x_d$, and velocity, $\ve_d$, in the world frame $\pazocal{F}_w$ can then be calculated from $^r\x_d$ and $^r\ve_d$ by employing conventional space transformations.

\subsection{Desired Position and Velocity as the Augmented States}
\label{subsec:desiredposandvelasaugmentedstates}
In the tracking controller of the end-effector in \cite{sleiman2021unified}, only the desired position is sent to the MPC planner as a fixed target with an interpolated trajectory from the current position, so the executed policy is still chasing the old target taken at the start of the MPC call. As discussed in Fig.~\ref{fig:telefail}, before the new policy is calculated, any new command from the operator will not affect the desired position of the end-effector, which is the reason for a significant delay and instabilities in the teleoperation loop. 

Knowing the desired end-effector velocity in addition to the target end-effector position, an alternative idea would be to send a trajectory rolled out by the desired velocity from the latest target position. This integrated trajectory will give the MPC planner the ability to predict the desired position of the end-effector over the horizon. In this work, we follow this idea by augmenting the states with the desired end-effector position and velocity such that the MPC can automatically roll out the trajectory of desired positions. The predicted reference trajectory can also be calculated outside MPC and sent into MPC directly, which will be discussed more in Sec.~\ref{sec:results}.

The set of augmented states is defined as $\x_{aug} = (\x_d, \ve_d) \in \mathbb{R}^6$, where $\x_d$ is the desired position of the end-effector, and $\ve_d$ is the desired velocity of the end-effector. Their dynamics can be described using the following equations
\begin{equation}
\begin{cases}
    \dot{\x}_{d} = \ve_d \\ 
    \dot{\ve}_{d} = \0 
\end{cases}
\label{eq:refposveldynamics}
\end{equation}
where we assume that the target velocity of the end-effector does not change in an MPC cycle. By defining the new state vector as $\x' = (\h_{com}, \q_b, \q_j, \x_d, \ve_d)\in \mathbb{R}^{12+n_a+6}$, the flow map of dynamic system $\dot{\x'} = f'(\x', \u)$ can be defined as the collection \eqref{eq:centroidaldynamics}, \eqref{eq:basecoordynamics}, \eqref{eq:jointdynamics}, and \eqref{eq:refposveldynamics}. In the definition of the tracking cost in \eqref{eq:trackingerror}, we use $\x_d$ from the augmented states with prediction instead of using the reference. 

In the MPC planner, we assume that the desired velocity of the end-effector is constant, which means that the input sequence generated will drive the end-effector to move straight with a constant velocity. Since the actual command can change during the MPC call, the affine-structured feedback policy from MPC can be used to correct this mismatch. As described in \cite{grandia2019feedback}, the policy taking the full states as feedback requires techniques such as frequency shaping to be implemented on hardware. In this work, we use the feedback policy partially to avoid this issue, i.e. we use the feedback term only on the augmented states. 

With desired position and velocity as augmented states, the SLQ-MPC algorithm can output a feedback policy which can take the new target as a correction. In this way, the operator can affect the input when the MPC policy is evaluated in the whole-body controller and a new policy is not yet available. In comparison with adding feedback on the actual end-effector position to reduce the tracking cost, using the desired position and velocity as feedback can be understood to increase the tracking cost, such that the input can be adapted to track the new target. If we assume the unaugmented states are the same as the nominal optimal ones, we can decompose the feedback gain matrix as $K(t) = [K_{unaug}(t), K_{aug}(t)]$, and write the resulting feedback policy as
\begin{equation}
    \pi(\x_{aug}(t), t) = \Bar{\u}(t) + K_{aug}(t)(\x_{aug}(t) - \Bar{\x}_{aug}(t)), 
    \label{eq:feedback_term}
\end{equation} 
where $\x_{aug}(t)$ is the current target position and velocity given by the operator and $\Bar{\x}_{aug}(t)$ is the predicted augmented states. 

In fact, the feedback gain matrices are the linear approximations of the optimal input, so the evaluated input is valid in the vicinity of the optimal states and constraint satisfactory when accounting for the new commands. Besides, the feedback policy is evaluated in the whole-body controller at a higher rate, which is independent on the update rate of MPC planner. Then the policy's output is also tracked in whole-body controller to generate the optimal joint torques $\boldsymbol{\tau}^*_j$, joint velocities $\dot{\q}^*_j$ and joint positions $\q^*_j$ by formulating it as a hierarchical quadratic program (QP) and solving it as an optimization problem \cite{bellicoso2016perception}. 

\section{Results}
\label{sec:results}
To evaluate the proposed framework, a set of experiments have been performed on a quadrupedal mobile manipulator which is composed of the ANYmal D platform\footnote{https://www.anybotics.com/} equipped with a torque-controlled 6-DoF robotic arm. On the operator side, we use the Omega 6 haptic device\footnote{https://www.forcedimension.com/products/omega} which is capable of providing force feedback in the 3-translational directions. The orientation of the device is ignored. The MPC planner, whole-body controller and the state estimator are all running on the robot's onboard computer (Intel core i7 8850H, 6 cores, 12 threads). The MPC planner is running at 70 Hz with a time horizon $T = 1$ s, while the whole-body controller and the state-estimator are running at 400 Hz. The low-level controller is generating torque commands at 2.5 kHz. The haptic device also runs at an update rate of 400 Hz. The reader is referred to the video attached with this paper for a better understanding of the experimental procedure.

\begin{figure*}[ht]
\centering
    \subfloat[]{\includegraphics[trim=0 17 0 6,clip,width=1\linewidth]{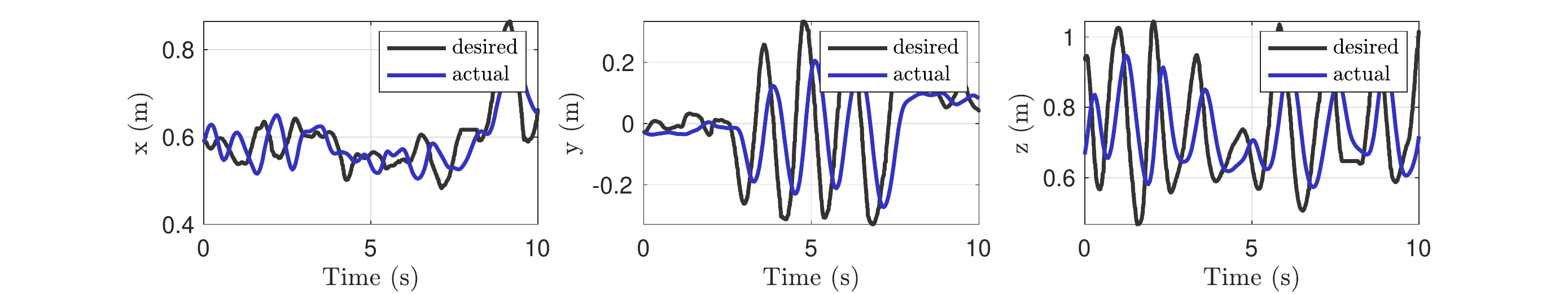}\label{fig:baseline_tracking}}
    
    \subfloat[]{\includegraphics[trim=0 17 0 6,clip,width=1\linewidth]{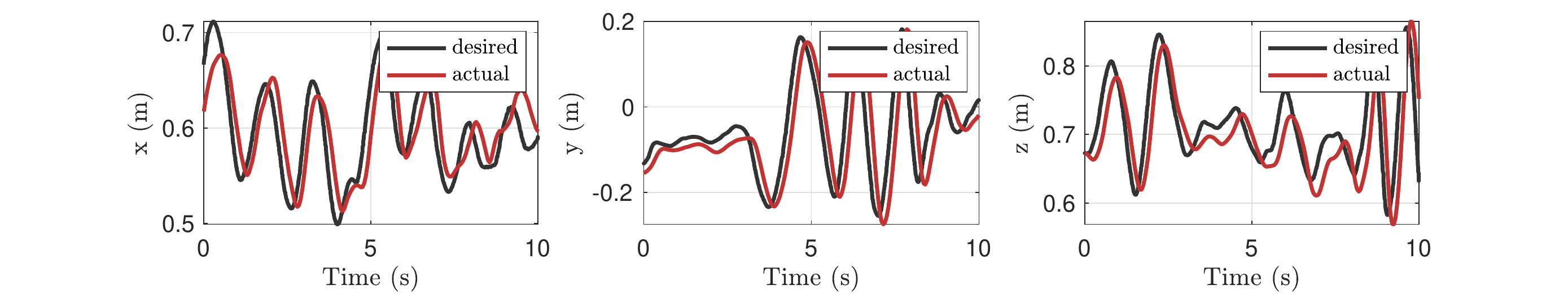}\label{fig:baseline_integral_tracking}}
    
    \subfloat[]{\includegraphics[trim=0 17 0 6,clip,width=1\linewidth]{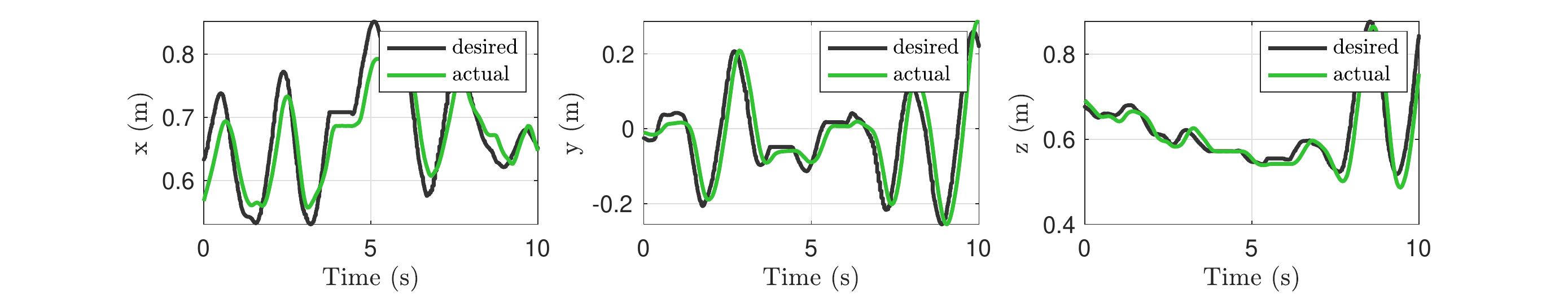}\label{fig:our_method_tracking}}
    \caption{Free motion tracking with the \protect\subref{fig:baseline_tracking} baseline controller,   \protect\subref{fig:baseline_integral_tracking} feed-forward MPC, and \protect\subref{fig:our_method_tracking}
    our method.}
    \label{fig:freemotion}
    \vspace{-3mm}
\end{figure*}


We mainly compared three different variants of the robot controller:
\begin{enumerate}
    \item \emph{Baseline}: The controller presented in Sec.~\ref{subsec:mpcplanner} is used. It takes as an input the current commanded position $\x_d$ and does not account for the velocity. The desired trajectory $\y_d$ is defined as a constant vector $\y_d = (\x_d, ..., \x_d)$.
    \item \emph{Feed-forward MPC}: The same baseline controller is used but we account for the commanded velocity $\ve_d$ by integrating it from the commanded position $\x_d$ over the MPC horizon such that $\y_d = (\x_d,..., \x_{d,i}, ..., \x_{d,n})$ where
    \begin{equation}
        \x_{d,i} = \x_d + \frac{i}{n} \ve_d T  
        \label{eq:baseline_with_integral}
    \end{equation}
    \item \emph{Our method}: The proposed framework described in Sec.~\ref{subsec:desiredposandvelasaugmentedstates} is used.
\end{enumerate}

The presented experiments focus on testing the following aspects which are essential for teleoperation:

\paragraph{Free motion tracking} The performance of free-motion tracking using the three mentioned methods is examined by moving the haptic device arbitrarily around. The trajectories of both the haptic device and the robot end-effector are recorded and compared.
    
\paragraph{Haptic feedback stability} The stability of the haptic feedback loop is evaluated. Six different tests are performed for each of the three different controllers to assess the stability of the haptic feedback loop. The operator here is asked to apply and maintain a constant force on a flat surface in front of the robot.
    
\paragraph{Constraint satisfaction} This experiment aims at verifying that the proposed feedback framework is constraint satisfactory within the vicinity of the MPC solution. Therefore, the robot end-effector is commanded to collide with the robot's body to check whether the self-collision avoidance constraint defined in \cite{chiu2022collisionfree} would be satisfied or not.

\subsection{Free Motion Tracking}
\label{subsec:freemotiontrackingusinghapticdevice}

\begin{figure}[t]
    \centering
    \centering
    \includegraphics[width=\linewidth]{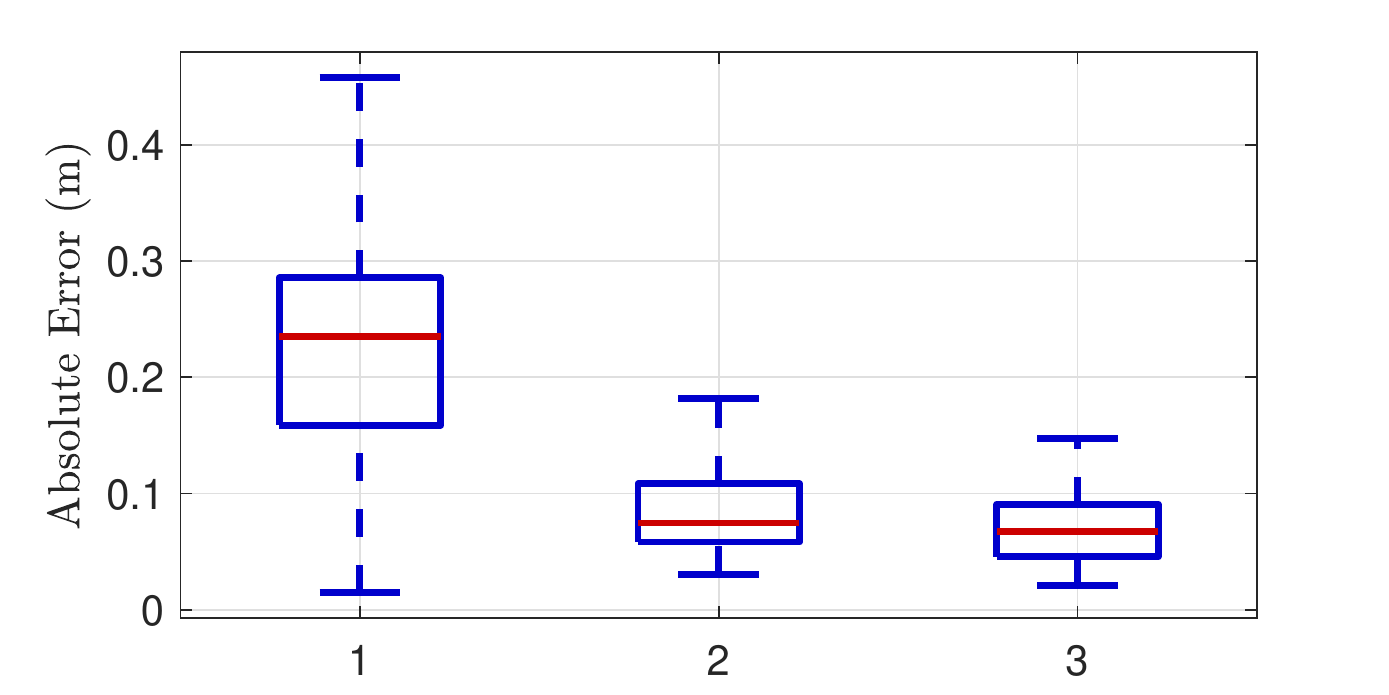}
    \caption{Free motion tracking error using the (1) baseline controller, (2) feed-forward MPC, and (3) our method.}
    \label{fig:freemotionerror}
    \vspace{-3mm}
\end{figure}

In this experiment, the operator acts on the haptic device to move the robot end-effector in free air. The trajectory followed by the robot is recorded and compared to the desired one sent from the haptic device. The results are depicted in Fig.~\ref{fig:freemotion} which shows the trajectory followed by the robot (in color) compared to the one commanded by the operator (in black) in the three translational directions. 

The results show that the baseline controller is slow in following the user commands (Fig.~\ref{fig:baseline_tracking}). In fact, this is expected since the controller is not provided with a feed-forward term informing about the target velocity and assumes that the target position is stationary. We rectify this in the feed-forward MPC controller where we account for the target velocity and provide the MPC with a predicted target trajectory integrated over the horizon as described in \eqref{eq:baseline_with_integral}. The qualitative tracking performance of the controller clearly improved as can be seen in Fig.~\ref{fig:baseline_integral_tracking}. A similar performance is observed with the method proposed in this paper where the robot manages to track the desired trajectory closely.


A quantitative analysis of the results is plotted in Fig.~\ref{fig:freemotionerror} which shows the absolute tracking error calculated by sampling over the traveled trajectories for each of the three different controllers. The results show that the median tracking error for the baseline controller is around $0.235~m$, that of feed-forward MPC is $0.075~m$ while the error of the method proposed in this paper is $0.067~m$.


\subsection{Haptic Feedback Stability}
\label{subsec:teleoperationwithforcefeedback}

\begin{figure}[t]
    \centering
    \subfloat[]{\includegraphics[trim=0 260 0 28,clip,width=1\linewidth]{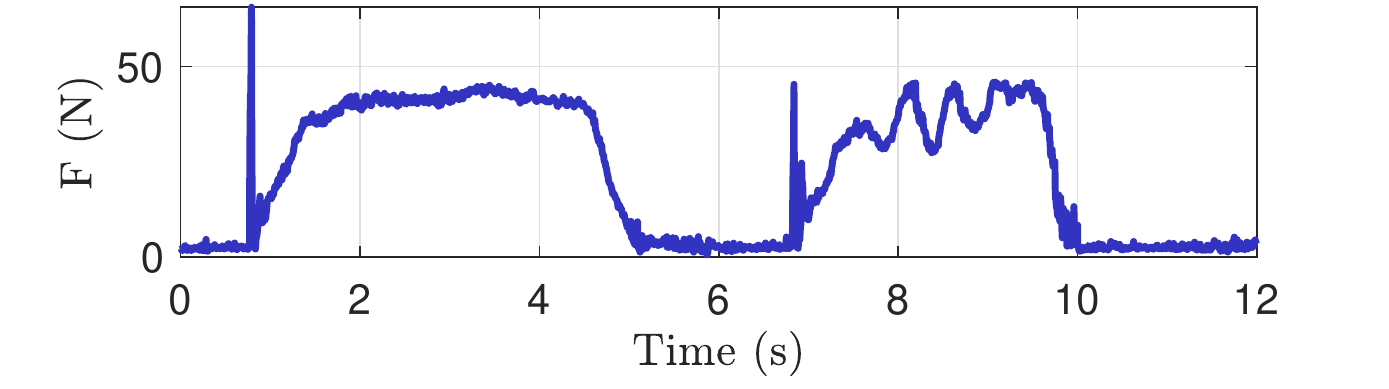}\label{fig:baseline_force}}
    \vspace{-1.5 mm}
    \subfloat[]{\includegraphics[trim=0 140 0 150,clip,width=1\linewidth]{figures/forcefeedback.pdf}\label{fig:baseline_integral_force}}
    \vspace{-1.5 mm}
    \subfloat[]{\includegraphics[trim=0 20 0 264,clip,width=1\linewidth]{figures/forcefeedback.pdf}\label{fig:our_method_force}}
    \vspace{-1.5 mm}
    \caption{The force feedback received by the user using \protect\subref{fig:baseline_force} the baseline controller, \protect\subref{fig:baseline_integral_force}
    feed-forward MPC and \protect\subref{fig:our_method_force} our method.
    }
    \label{fig:teleopff}
    \vspace{-5 mm}
\end{figure}

\begin{figure}[t]
    \centering
    \includegraphics[width=1\linewidth]{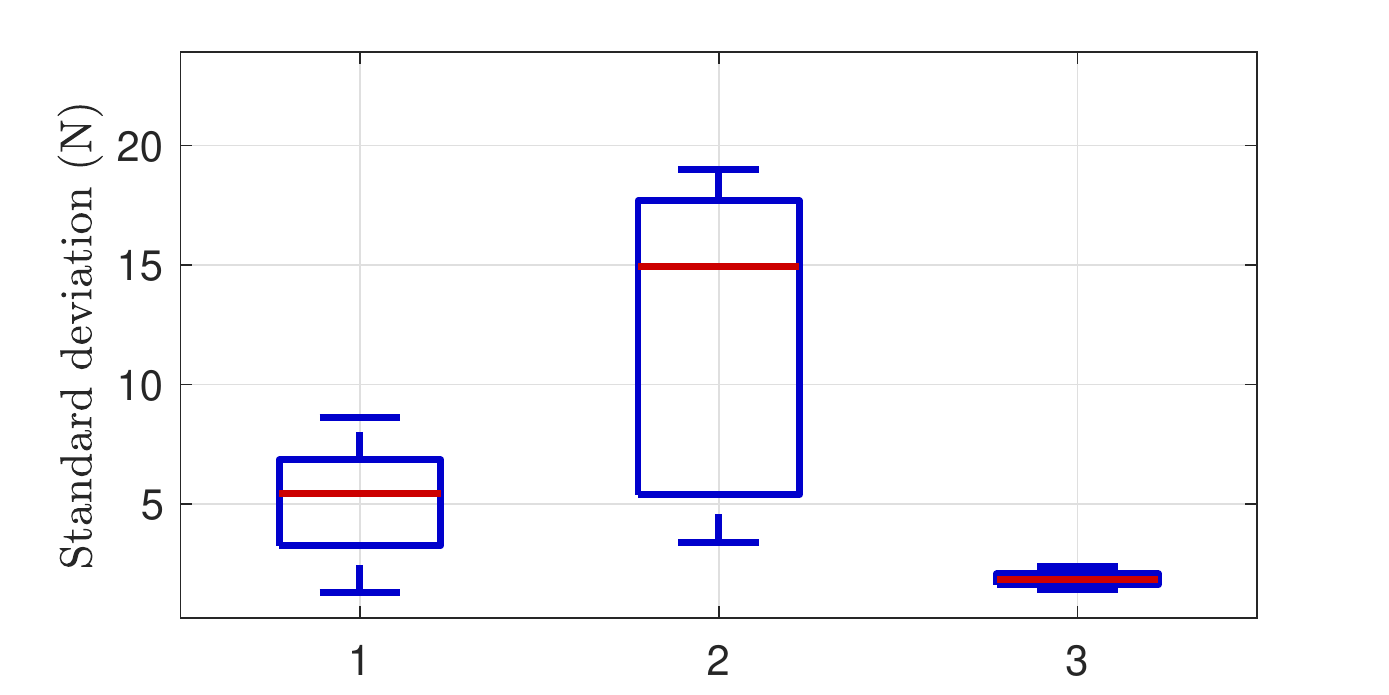}
    \caption{The standard deviation of force feedback over a period of 1.5~s after contact, using the (1) baseline controller, (2) feed-forward MPC, and (3) our method.}
    \label{fig:teleopfcom}
    \vspace{-5 mm}
\end{figure}

In the second experiment, the operator tries to command the end-effector to make contact with a solid wooden block placed in front of the robot (see Fig.~\ref{fig:experiment}) and maintain a constant force with the block. The experiment was repeated 18 times (6 for each of the three tested controllers). Fig.~\ref{fig:teleopff} shows the magnitude of the force fed back to the operator (read from a force torque sensor at the wrist of the robot) during two representative instances of the mentioned experiments for each of the three tested controllers. 


Using the baseline controller, the user experienced relatively low vibrations. In some cases mild vibrations were observed (see Fig.~\ref{fig:baseline_force}), but the system remained stable. In fact, the MPC costs were tuned to have a compliant end-effector behavior for the three controllers. This compliance helped ensure the stability of the haptic loop even with the slow update rate of the baseline controller. On the other hand, the feed-forward MPC controller drove the system unstable as soon as the operator tried to establish contact with the surface as shown in Fig.~\ref{fig:baseline_integral_force}. This is expected as human reflexes during haptic interaction are highly discontinuous and the assumption that the commanded velocity (which is provided as a feed-forward term to the MPC in this case) will remain constant over the horizon is not valid. Finally, Fig.~\ref{fig:our_method_force} reports the results obtained using the proposed method. The haptic loop remained stable in all contact instances even with high impact contacts where the initial bump force jumped up to more than $70~N$. This, of course, can be clearly explained by the much faster update rate of the controller which accounts for new commands with a minimal delay.

Finally, a quantitative statistical analysis of the results obtained during this experiment is performed by estimating the average standard deviation of the force feedback received by the operator over a period of 1.5 seconds after contact was established with the environment (we assumed the contact was established 0.1 seconds after the initial bump). The results reported in Fig.~\ref{fig:teleopfcom} demonstrate that the proposed framework clearly outperformed both the baseline and the feed-forward MPC in terms of the stability of the haptic feedback loop. Note that while a passivity enforcement algorithm could be used to damp the vibrations observed in the feed-forward MPC and the baseline controllers to ensure the stability of the haptic feedback loop, we did not use such an algorithm in order to clearly demonstrate the performance of the controllers themselves and their impact on the bilateral teleoperation experience. 


\subsection{Constraint Satisfaction}
\label{subsec:constraintsatisfactionusingourmethod}

\begin{figure}[t]
    \centering
    \centering
    \includegraphics[trim=0 0 0 10,clip,width=\linewidth]{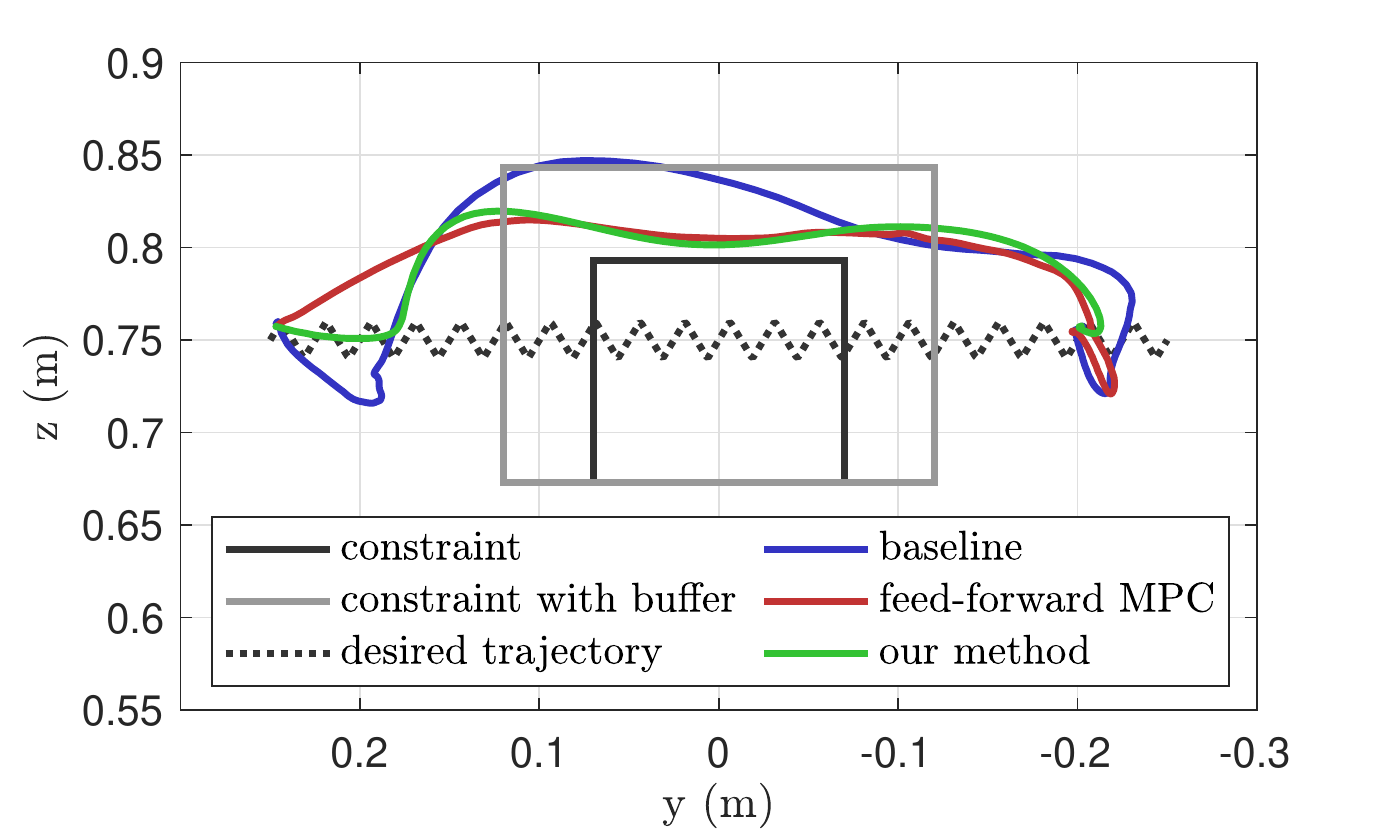}
    \caption{Constraint satisfaction validation test: the robot is commanded to collide with itself. The collision body is shown as the black constraint with a soft buffer shown in grey. The robot is commanded over the dotted trajectory and the executed trajectories using the different controllers are shown in color.}
    \label{fig:constraint}
    \vspace{-5mm}
\end{figure}

A key strength of MPC-based control schemes is their ability to account for various constraints. This is especially important for a legged manipulator where several constraints should be respected to ensure successful task completion. In this experiment, we seek to validate that the proposed framework satisfies the constraints defined in the non-linear optimal control problem described in \ref{subsec:mpcplanner} even with the feedback term introduced in \eqref{eq:feedback_term}. To this end, we command the robot on a trajectory to collide with its own body. In particular, the end effector is commanded over a trajectory which goes through a LiDAR on the back of the robot. The LiDAR and its cage are modeled as a cylinder which is shown in Fig.~\ref{fig:constraint} from a front-end view.
In the figure, the LiDAR constraint is plotted as a rectangle in 2D. The end effector is commanded over an oscillatory zig-zag trajectory starting from the left side to the right side. The trajectory is chosen to mimic the human hand motion which is never noise-free and to ensure that the feedback term is indeed having an effect on the motion. 
The experiment shows that even with such a challenging oscillatory commanded trajectory, our proposed method respects the constraint and avoids collision by going above the LiDAR cage. 

\section{Discussions and Conclusions}
\label{sec:conclusion}
We present a novel framework for teleoperating complex robotic systems within an MPC scheme. We build on recent contributions on whole-body feedback MPC controllers \cite{grandia2019feedback} and employ their structure to account for the operator input at a high rate with the goal of improving the stability and transparency of the bilateral teleoperation loop. The proposed architecture still respects the constraints defined in the MPC optimal control problem by considering them in the feedback-gain matrices. Finally, hardware experiments were performed on a quadrupedal manipulator such that the proposed method was tested against the classic MPC controller, and a feed-forward MPC controller, which only predicts the evolution of the user command over the horizon without introducing feedback. The experiments focused on three aspects: 1) tracking accuracy in free motion, 2) haptic-loop stability during interactions with the environment, and 3) system constraints. 

The primary motivation behind this work is that the state-of-the-art, whole-body MPC formulation is an order of magnitude slower than the frequency required for stable bilateral teleoperation. This was reflected in the free motion experiment where the end-effector lagged behind the user commands. This issue could be solved by integrating a feed-forward term in the MPC formulation, which assumes that the velocity command of the user is constant within one controller iteration. However, it generated significant instabilities during haptic interactions with the environment. On the other hand, the proposed framework showed excellent performance in both scenarios and proved to be a viable solution for bilateral teleoperation within an MPC framework. The final experiment also showed that the framework does not violate the constraints defined in the optimal control problem.


A possible extension of the proposed framework is to control the orientation of the robot end-effector in addition to its position by adding the target orientation and angular velocity as additional augmented states. It would go hand-in-hand with using a haptic device capable of providing force feedback on all six degrees of freedom instead of just three, as is the current case in the used haptic device. 
Besides, our method assumes that the target accelerations are zero in the augmented model of target velocities, which might cause overshooting when the actual target accelerations are large. Therefore, it could be interesting to explore the effect of accounting for target accelerations into the MPC formulation as augmented states as well.

\bibliographystyle{IEEEtran}
\bibliography{shared}

\end{document}